\documentclass[review]{elsarticle}

\usepackage{lineno,hyperref}
\usepackage{graphicx}
\usepackage{color}
\usepackage[marginal]{footmisc}
\usepackage{subfigure}
\usepackage{amsmath}
\usepackage[noend]{algpseudocode}
\usepackage{algorithmicx}
\usepackage{algorithm}
\usepackage{setspace}
\usepackage[justification=centering]{caption}
\usepackage{multirow}
\usepackage{booktabs}
\usepackage{verbatim}

\journal{Journal of Knowledge-Based Systems}

\begin{document}

\begin{frontmatter}

\title{NROWAN-DQN: A Stable Noisy Network with Noise Reduction and Online Weight Adjustment for Exploration}

\author[mymainaddress,mysecondaryaddress]{Shuai Han}

\author[mymainaddress,mysecondaryaddress]{Wenbo Zhou}
\author[mymainaddress,address3]{Jing Liu}

\author[mymainaddress,mysecondaryaddress]{Shuai L\"u\corref{mycorrespondingauthor}}
\cortext[mycorrespondingauthor]{Corresponding author}
\ead{lus@jlu.edu.cn}




\address[mymainaddress]{Key Laboratory of Symbolic Computation and Knowledge Engineering (Jilin University), Ministry of Education, Changchun 130012, China}
\address[mysecondaryaddress]{College of Computer Science and Technology, Jilin University, Changchun 130012, China}
\address[address3]{College of Software, Jilin University, Changchun 130012, China}

\begin{abstract}
Deep reinforcement learning has been applied more and more widely nowadays, especially in various complex control tasks. Effective exploration for noisy networks is one of the most important issues in deep reinforcement learning. Noisy networks tend to produce stable outputs for agents. However, this tendency is not always enough to find a stable policy for an agent, which decreases efficiency and stability during the learning process. Based on NoisyNets, this paper proposes an algorithm called NROWAN-DQN, i.e., Noise Reduction and Online Weight Adjustment NoisyNet-DQN. Firstly, we develop a novel noise reduction method for NoisyNet-DQN to make the agent perform stable actions. Secondly, we design an online weight adjustment strategy for noise reduction, which improves stable performance and gets higher scores for the agent. Finally, we evaluate this algorithm in four standard domains and analyze properties of hyper-parameters. Our results show that NROWAN-DQN outperforms prior algorithms in all these domains. In addition, NROWAN-DQN also shows better stability. The variance of the NROWAN-DQN score is significantly reduced, especially in some action-sensitive environments. This means that in some environments where high stability is required, NROWAN-DQN will be more appropriate than NoisyNets-DQN.	
\end{abstract}

\begin{keyword}
\texttt{}Deep reinforcement learning\sep Exploration\sep Noisy networks\sep Noise reduction\sep Online weight adjustment
\end{keyword}

\end{frontmatter}

\section{Introduction}

Deep reinforcement learning has been successfully applied in various complex control tasks, such as robotic control tasks \cite{Ref20166} \cite{Ref20178} \cite{Ref20154} \cite{Ref20179}, games \cite{Ref20151} \cite{Ref20163} \cite{Ref20181} \cite{Ref20167}, natural language processing \cite{Ref201710} \cite{Ref20183} and recommendation systems \cite{Ref20156}. Deep Q-network (DQN) \cite{Ref20151} is one of the most widely used deep reinforcement learning algorithms. However, there still exist a lot of problems about efficiency and stability of the original DQN, such as learning disability in sparse reward environments \cite{Ref20177} and exponential training time in delayed reward environments \cite{Ref20165}. A considerable part of these problems lies in immaturity of the exploration mechanism. Exploration is considered as a key challenge in reinforcement learning \cite{Ref20171}. 

When it comes to guiding an agent to interact with environments, the simplest exploration method is dithering actions by random factors, such as $\varepsilon$-greedy \cite{Ref1998}. However, when an action-state space is large, this way of exploration may be inefficient. After making some optimistic assumptions, heuristic exploration can provide theoretical guarantees for agent performance \cite{Ref20172}. But this approach is usually limited to small action-state spaces \cite{Ref20173}. Some efficient exploration methods have been proposed in recent years. Tang et al. extended the classical counting method in reinforcement learning to high-dimensional spaces by using counting tables, but their additional components are complex  \cite{Ref20174}; Houthooft et al. presented a practical implementation using variational inference in Bayesian neural networks, which efficiently handles continuous state space and action space \cite{Ref20161}. However, this method dynamically modifies rewards of environment, resulting in the MDP process non-stationary to the agent. As a novel approach, object-oriented Q-map agent \cite{Ref20182} conduct effective exploration by disturbing target instead of action so that action in the exploration process are more consistent. Recent experiments have shown that adding noises in the parameter domain rather than the action domain can lead to better exploration \cite{Ref20175}. Based on this principle, Fortunato et al. \cite{Ref20173} applied noisy networks to DQN, A3C \cite{Ref20163} and DuelingNet \cite{Ref20152} algorithms, and developed NoisyNet-DQN, NoisyNet-A3C and NoisyNet-Dueling, which got higher scores in Atari 2600. For convenience, we use NoisyNets to generally refer to NoisyNet-DQN, NoisyNet-A3C and NoisyNet-Dueling.

\begin{figure}[h]
	\centering
	\subfigure[Action domain noise in DQN and its action noise curve]{
		\begin{minipage}[t]{1\linewidth}
			\centering
			\includegraphics[width=4.8in]{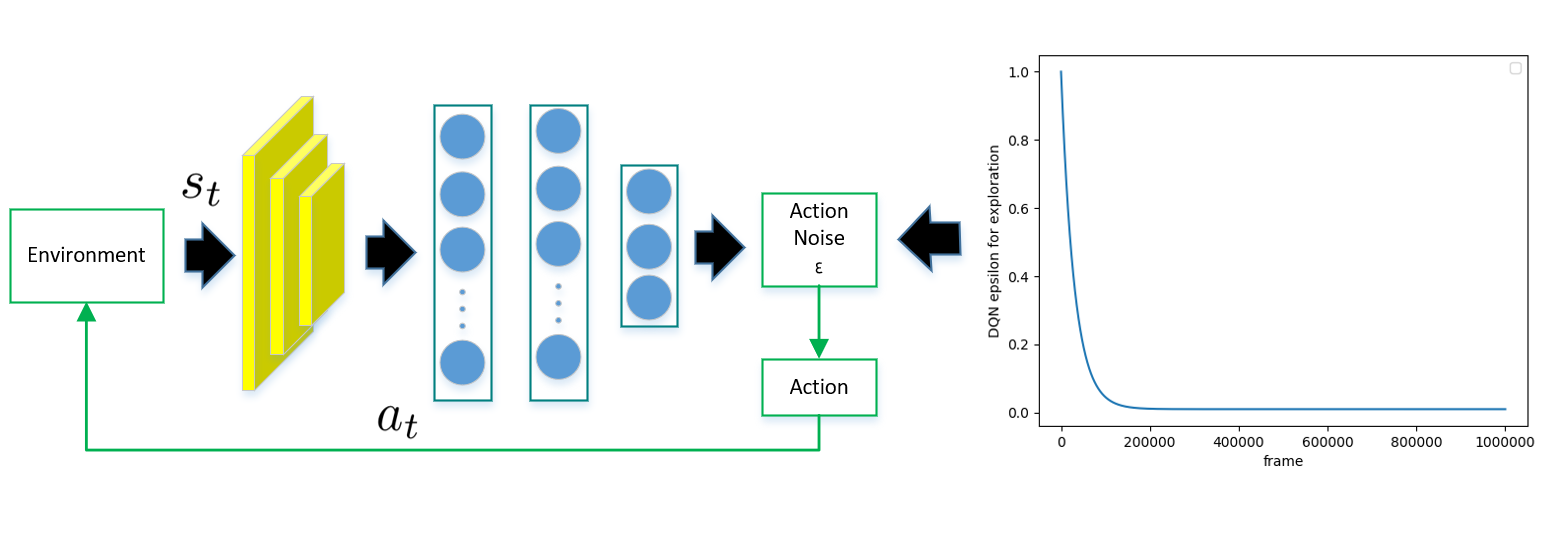}
			\label{fig:DQN}
		\end{minipage}%
	}%
	
	\subfigure[Parameter domain noise in NoisyNet-DQN and its action noise curve]{
		\begin{minipage}[t]{1\linewidth}
			\centering
			\includegraphics[width=4.8in]{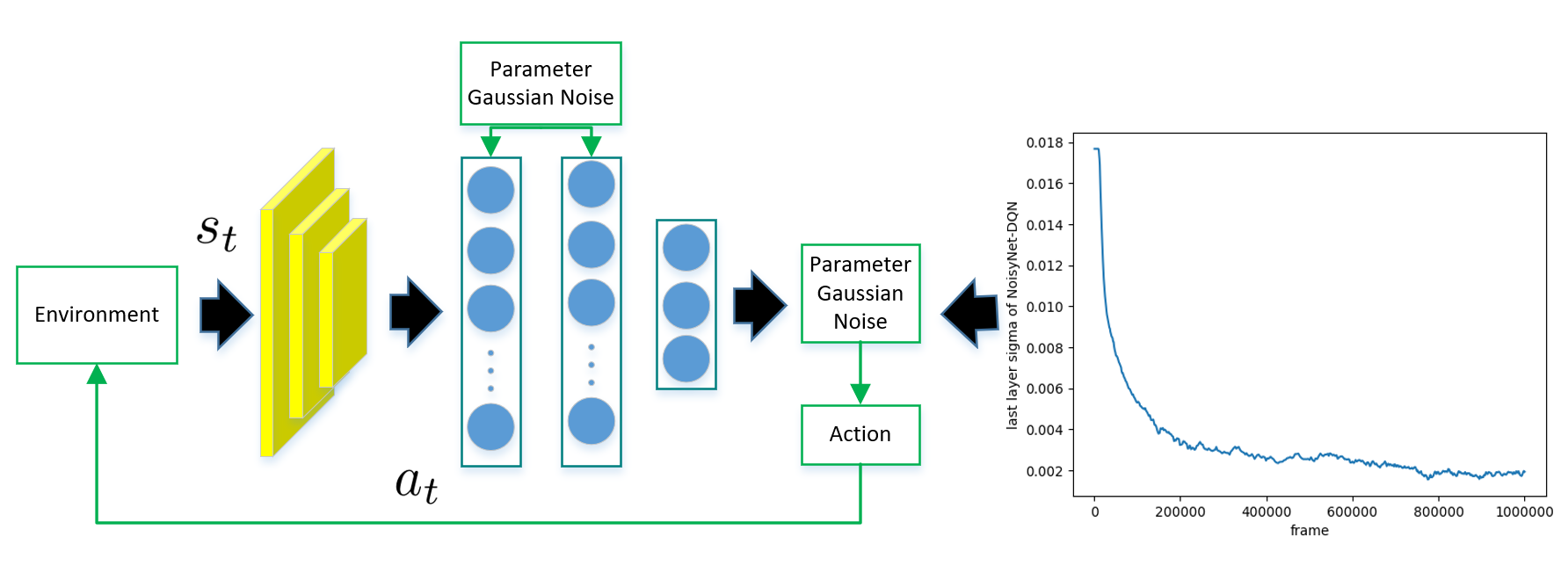}
			\label{fig:NNDQN}
		\end{minipage}%
	}%
	\centering
	\caption{Different noise methods and action noise curves}
\end{figure}

The excellent performance of NoisyNets is mainly because networks with noisy parameters bring more abundant exploration. With these fully explored samples, an agent is more likely to jump out of local optimum when learning an optimal policy. However, noisy parameters also limit the efficiency of algorithms. In Fortunato's model \cite{Ref20173}, the reduction of noise spontaneously occurrs with a random gradient descent of TD-error, which results in a slow and insufficient reduction of noise during the learning process. Figure \ref{fig:DQN} shows the decision-making process of the original DQN and the change of its action noise. In a classical DQN setting, the action noise level ($\varepsilon$ value) decreases to 1\% of the initial value at 15K frames, and then the agent interacts with the environment with a relatively stable action policy. However, as shown in Figure \ref{fig:NNDQN}, the noise level decreases to only 22\% of the initial value at 15K frames with the parameter domain noise setting in NoisyNet-DQN. The slow and insufficient spontaneous decrease of $\sigma$ makes it difficult to form a stable policy quickly, which affects performance of the agent.

In order to solve the problem that NoisyNets cannot form a stable policy effectively, this paper designs a differentiable online noise reduction mechanism, which can help agents form stable action policy based on parameter domain exploration. The core of this noise reduction mechanism is a deterministic factor which is differentiable to noise parameters, so this mechanism can be perfectly combined with the learning process. In the experimental part, we demonstrate that NROWAN-DQN has higher scores and better stability than the previous algorithms in both low-dimensional space and high-dimensional space. With the better stability, NROWAN-DQN is more practical than NoisyNet-DQN in some action-sensitive environments. Finally, we further explore relationship between the new parameter and the learning rate, and how they affect NROWAN-DQN performance.

The rest of this paper is organized as follows: Section 2 introduces some related work; Section 3 provides basic background; Section 4 proposes a noise reduction mechanism, an online weight adjustment strategy, and a NROWAN-DQN algorithm; Section 5 presents parameters setting and experimental evaluation of NROWAN-DQN; Section 6 concludes the paper and outlines the future work.

\section{Related Work}

This paper involves the exploration field of reinforcement learning. Currently, the methods to solve the exploration problems mainly include heuristic algorithm, state-space modeling, curiosity mechanism, and stochastic method.

\textbf{Heuristic algorithm.} Heuristic exploration can provide theoretical guarantees for agent performance. Thomas et al. proposed UCRL2, which achieves a (gap-dependent) regret bound that is logarithmic in learning step $T$ \cite{Ref20101}; With some optimistic assumptions, Mohammad et al. proposed UCBVI-BF, which applies a conncentration to the value as a whole and a recursive law of total variance to couple estimates across an episode. This has improved over the bound achieved by the UCRL2 algorithm \cite{Ref20172}; Christoph et al. considered Episodic Fixed-Horizon MDPs, and proposed UCFH, which improves on previous bounds for episodic finite-horizon MDPs \cite{Ref20157}.

\textbf{State-space modeling.} State-space modeling helps an agent to efficiently find states with more information. There are two ways to model state space. One way is the count-based exploration: Strehl et al. proposed a variation of Model-based Interval Estimation called MBIE-EB \cite{Ref20081}; Haoran et al. mapped states to hash codes, accordingly applying count-based methods in high-dimensional state spaces \cite{Ref20174}; Marc et al. proposed an algorithm for deriving a pseudo-count from an arbitrary density model, which can generalize count-based exploration algorithms to the non-tabular case \cite{Ref20162}. The other way is called states density modeling. These methods usually drive exploration by estimating state space density \cite{Ref201711} \cite{Ref201712} \cite{Ref20191}.

\textbf{Curiosity mechanism.} One kind of curiosity mechanism is generated based on the unknown information of the environment, such as Bayesian exploration in an unknown dynamic environment \cite{Ref20111} and the theory based on the concept of maximizing intrinsic reward for active creation or discovery of new states \cite{Ref20102}. The other kind of curiosity mechanism focuses on prediction of future state, such as forming intrinsic rewards that approximate the KL-divergence of true transition probabilities from a learned model \cite{Ref20177} and to formulate curiosity as the error in an agent's ability to predict the consequence of its own actions in a visual feature space learned by a self-supervised inverse dynamics model \cite{Ref201713}.

\textbf{Stochastic method.} Adding noises to the parameter domain rather than the action domain can lead to better exploration \cite{Ref20175}. Based on this principle, Fortunato \cite{Ref20173} combined noisy networks with DQN, A3C \cite{Ref20163} and DuelingNet \cite{Ref20152}; Plappert et al. introduced parameter space noise to DQN, DDPG \cite{Ref20154} and TRPO \cite{Ref20155}. Both of the above approaches deliver excellent performance.

There are also some other novel methods about exploration. Houthooft et al. proposed an exploration strategy based on variational information maximization \cite{Ref20161}; Xiong et al. and Adam et al. developed some reason-based exploration strategies, respectively \cite{Ref201714} \cite{Ref20184}; Target-based short trajectory exploration is also one of the novel methods \cite{Ref20182} \cite{Ref20185}. These methods conduct effective exploration by disturbing target instead of disturbing action.

Our method belongs to stochastic methodes for exploration. Based on NoisyNets, this paper advocates limiting noise level of the last layer of NoisyNets, and proposes the NROWAN-DQN algorithm. More specifically, this paper designs a differentiable online noise reduction mechanism. This mechanism can help agents form stable action policies based on parameter domain exploration. In addition, because the deterministic factor for reducing noise is differentiable, this mechanism can be perfectly combined with the learning process.

\section{Background}

\label{sec:1}
As a preparation for our method, this section introduces Markov decision process, DQN and NoisyNet-DQN.
\subsection{Markov decision process in reinforcement learning}
\label{sec:2}
One of the main tasks in reinforcement learning is to solve the problem that how an agent learns to take actions to maximize rewards during the interaction with environments. There is no direct supervision in learning process. For example, an agent never knows what the optimal action is \cite{Ref20181}. The interaction between agent and environment can be abstracted into Markov decision processes.

At each discrete time step $t$ (i.e., $t$ = 0, 1, 2, $\cdots$), the environment provides a state \begin{math} s_t  \end{math} to an agent, and the agent takes an action \begin{math} a_t \end{math} as a response to this observation. Then, the environment returns a reward \begin{math}r_{t}\end{math}, a discount factor \begin{math} {\gamma}_{t+1}  \end{math}, and the next state \begin{math} s_{t+1}  \end{math}. The interaction process at each time step is denoted as a 5-tuple \begin{math} (S,A,P,R,\gamma) \end{math}, where \begin{math} S  \end{math} is a finite set of states, \begin{math} A  \end{math} is a finite set of actions, \begin{math} P(s_{t+1}|s_t,a_t)  \end{math} is a state transition probability function, \begin{math} R  \end{math} is a reward function, \begin{math} \gamma\in[0,1]  \end{math} is a discount factor. In the experimental part of this paper, the discount factor is set to a constant in accordance with most other experiments.

In reinforcement learning, an agent selects actions according to its policy. A policy \begin{math} \pi(a_t|s_t)  \end{math} is the conditional probability distribution of \begin{math} a_t \end{math} at state \begin{math} s_t  \end{math}. For any observation, the total reward of an agent at its current state is defined as follow:
$$ G_t = \sum_{k=0}^{\infty}\gamma_t^kR_{t+k+1}\eqno(1) $$
where \begin{math} \gamma_t^kR_{t+k+1} \end{math} represents the reward obtained at the next step \begin{math} k \end{math}. The goal of the agent is to find an optimal policy by maximizing its discount total reward. In some cases, a policy can be obtained directly. However, sometimes a policy needs to be represented as a parametric equation that can be solved using a learning algorithm. In Q-value based reinforcement learning, an agent learns a Q function, which is denoted as:
$$ Q^\pi(s,a) = E^\pi(G_t|s_t=s,a_t=a)\eqno(2) $$

The way to iterate a new policy from a state action function is \begin{math} \varepsilon\end{math}-greedy. In detail, at each discrete time step, an random action is taken with the probability \begin{math} \varepsilon\end{math}, or a greedy action is taken with the probability $1-\varepsilon$ to maximize \begin{math} Q^\pi \end{math}.

\subsection{Deep reinforcement learning and DQN}
\label{sec:3}
When the state or action space is large, it is very difficult for an agent to directly learn a value function or a policy. We usually use deep neural network to approximate a value function or a policy. The former is called value based deep reinforcement learning, while the latter is called policy gradient based deep reinforcement learning.

DQN \cite{Ref20151} is a typical value based reinforcement learning algorithm. It uses TD-error as the loss function of a neural network, and moreover, it involves convolution, experience buffer and random experience replay. DQN successfully got humen-level scores in the Atari domain. At each time step, the environment provides an agent with an observation \begin{math} s_t  \end{math}. First, the agent takes action according to $\varepsilon$-greedy strategy, and receives the environment response \begin{math}r_{t}\end{math} and \begin{math} s_{t+1}  \end{math}. Then, a 4-tuple \begin{math} (s_t,a_t,r_{t},s_{t+1}) \end{math} is pushed into a experience buffer. Finally, the algorithm takes samples from the experience buffer, and uses random gradient descent to minimize the loss function.
$$ 
L(\theta) = E[r_{t}+\gamma \underset{a'}{max}Q^\pi(s_{t+1},a';\theta^-)-Q^\pi(s_t,a_t;\theta)]
\eqno(3) $$
where \begin{math} \theta  \end{math} is the parameter of online network and \begin{math} \theta^-  \end{math} is the parameter of target network. \begin{math} \theta^-  \end{math} is updated to \begin{math} \theta  \end{math} at every certain interval.

\subsection{DQN with noisy networks}
\label{sec:4}
Noisy networks \cite{Ref20173} refers to a neural network in which both weights and biases are disturbed by noise. We use \begin{math} Q^\pi(s,a;\theta) \end{math} to denote a parameterized \begin{math} Q \end{math} function. When using noisy networks to parameterize \begin{math} Q \end{math} functions, \begin{math} \theta \stackrel{def}= \mu + \Sigma\odot\varepsilon \end{math}. We use \begin{math} \zeta \stackrel{def}= (\mu, \Sigma) \end{math} to denote a learnable parameter during learning, $\varepsilon$ denotes a zero-mean random vector with fixed statistics, and $\odot$ denotes the element-wise multiplication between vectors.

Specifically, a fully connected layer with a $p$-dimension input and a $q$-dimension output in a neural network can be written as \begin{math} y = w\cdot{x}+b \end{math}. The corresponding noisy layer is defined as:
$$ y = (\mu^w+\sigma^w\odot{\varepsilon^w})\cdot{x}+(\mu^b+\sigma^b\odot{\varepsilon^b}) \eqno(4) $$
where \begin{math} \mu^w, \sigma^w \in R^{q\times{p}}, \mu^b , \sigma^b \in R^q \end{math} are learnable parameters, and $ \varepsilon^w \in R^{q\times{p}},$ $  \varepsilon^b \in R^q $ are noises. Figure 2 shows a classical linear layer and a noise linear layer, respectively. In the experimental part, we use Factorised Gaussian noise to generate $p$ independent Gaussian noises \begin{math} \varepsilon_i \end{math} and $q$ independent Gaussian noises \begin{math} \varepsilon_j \end{math}. Then, every single weight noise \begin{math} \varepsilon_{i,j}^w \end{math} and bias noise \begin{math} \varepsilon_{j}^b \end{math} can be calculated as follows: 
$$  \varepsilon_{i,j}^w = sgn(\varepsilon_i\cdot{\varepsilon_j})\cdot\sqrt{|\varepsilon_i\cdot\varepsilon_j|} \eqno(5) $$
$$  \varepsilon_j^b = sgn(\varepsilon_j)\cdot\sqrt{|\varepsilon_j|} \eqno(6) $$

\begin{figure}[h]
	\begin{spacing}{0.7}	
		\centering
		\subfigure[A linear layer]{
			\begin{minipage}[t]{0.5\linewidth}
				\centering
				\includegraphics[width=2in]{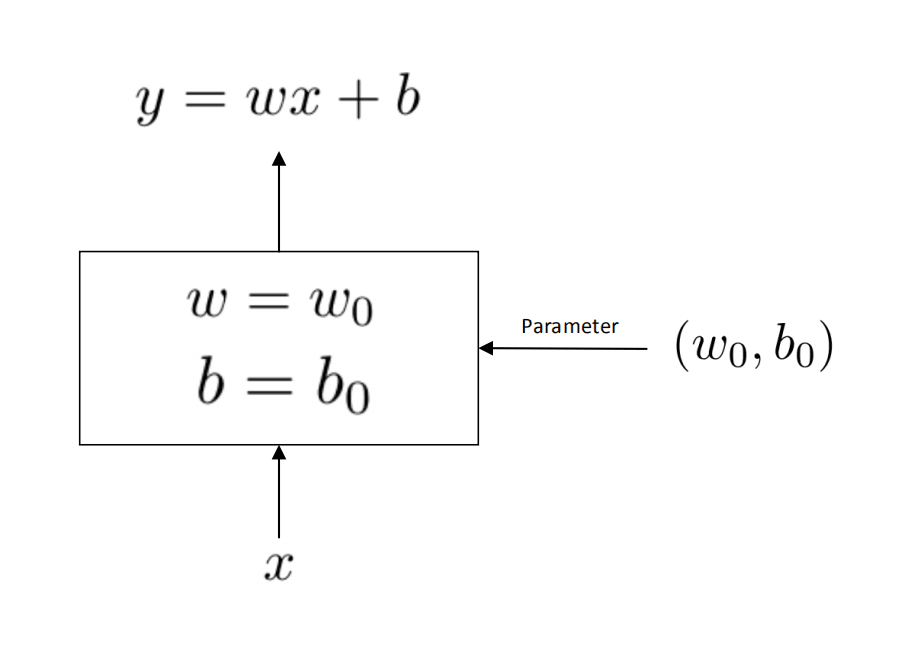}
			\end{minipage}%
		}%
		\subfigure[A noisy linear layer]{
			\begin{minipage}[t]{0.5\linewidth}
				\centering
				\includegraphics[width=2in]{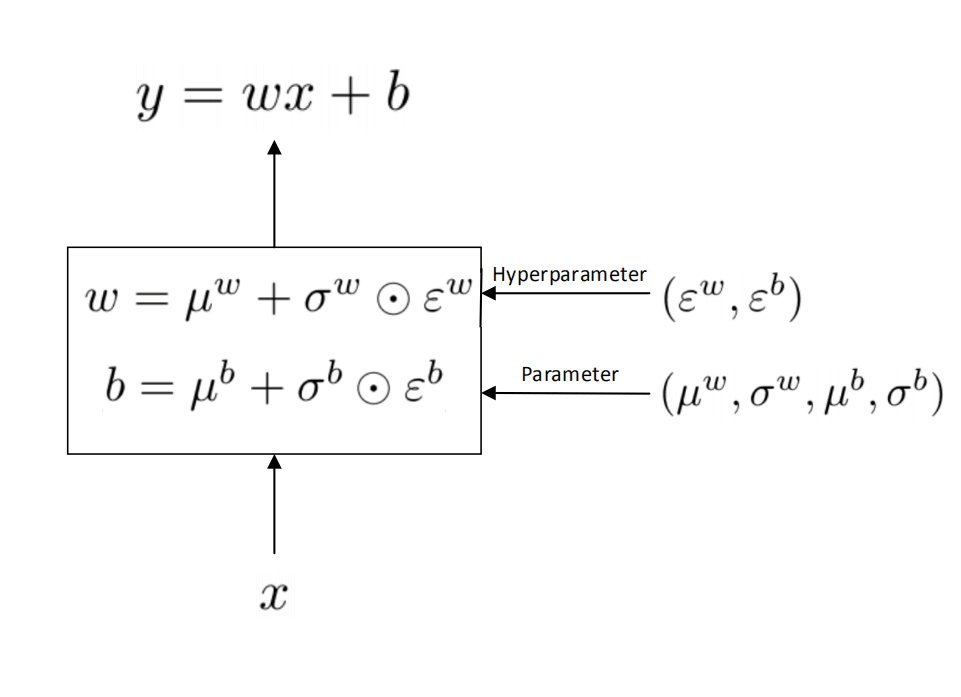}
			\end{minipage}%
		}%
		\centering
		\caption{Graphical representation of a linear layer and a noisy linear layer}	
	\end{spacing}
\end{figure}

Since loss of a noisy network is denoted as an expectation over noise, the gradient can be obtained using \begin{math} \hat{L}(\zeta) = E[L(\theta)] \end{math}. The parameters in the original DQN are replaced with learnable parameters in noisy networks:
$$ \hat{L}(\zeta) = E[E[r_{t}+\gamma \underset{a'}{max}Q^\pi(s_{t+1},a',\varepsilon;\zeta^-)-Q^\pi(s_t,a_t,\varepsilon;\zeta)]] \eqno(7) $$

\section{Online noise reduction for noisy networks}
This section presents two main mechanisms of NROWAN-DQN, including noise reduction and online weight adjustment.

\subsection{Noise reduction}

\begin{figure}[h]
	\centering  
	\includegraphics[width=0.75\linewidth]{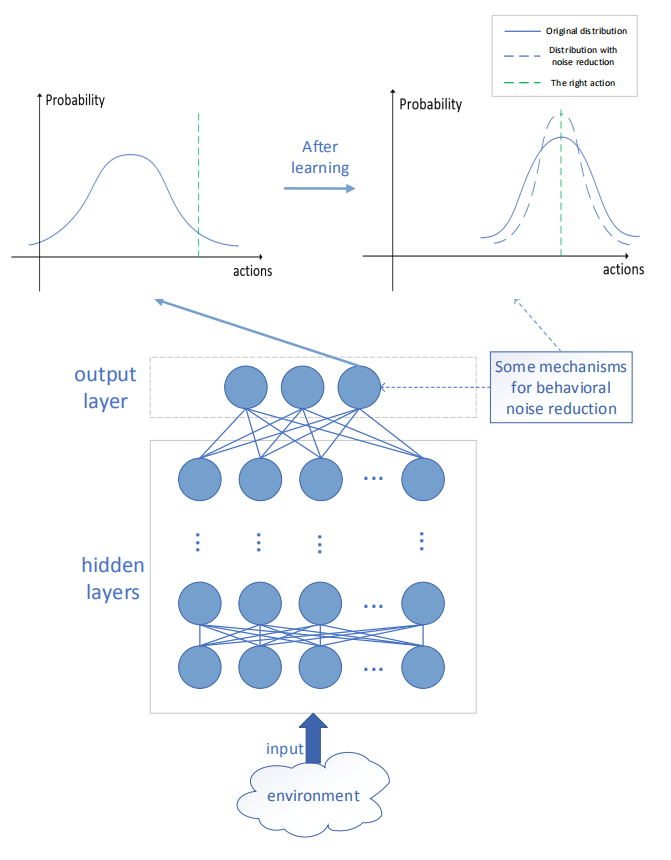}  
	\caption{The scope and effect of the noise reduction mechanism in NoisyNet-DQN}  
	\label{fig:framework}   
\end{figure}

The instability of NoisyNets during the learning process is mainly affected by the noise variance \begin{math} \sigma \end{math}, so the noise can be reduced by decreasing \begin{math} \sigma \end{math}. However, Fortunato et al. pointed out that in some environments, \begin{math} \sigma \end{math} of Noisy Networks' hidden layer may increase with the progress of learning, and its value maintains a large value after an agent forms a stable policy \cite{Ref20173}. This indicates that a larger \begin{math} \sigma \end{math} in Noisy Networks' hidden layer has a positive effect. And because the noise variance of output layer directly affects the noise of actions, we limit the overall noise level of the Q network by controlling the \begin{math} \sigma \end{math} of the output layer rather than that of all layers. The scope and effect of noise reduction mechanism are shown in Figure \ref{fig:framework}. The output probability of each neuron in the last layer is a normal distribution (blue line). At this time, the probability that this neuron output a correct action (green dotted line) is small. After sufficient learning, the mean of the distribution that the neuron output should be consistent with the correct action. That is, the neuron should be able to select the correct value with a higher probability after sufficient learning. According to the noise reduction mechanism, the distribution learned by neurons (blue dotted line) should have a smaller variance. Therefore these neurons have a greater probability of choosing the right action.

A general scheme for noise attenuation is to gradually reduce the \begin{math} \sigma \end{math} of the output layer during learning progress. However, in noisy networks, the \begin{math} \sigma \end{math} of the output layer is a parameter that is updated along the gradient direction. Therefore, reducing its value independently may cause this parameter not to be updated in gradient direction. This may result in inefficiency of learning and even prevent the agent from learning a valid policy. Therefore, we need a noise reduction mechanism that is consistent with learning process.

Our idea is to represent the noise level in a differentiable form. We use $D$ to denote the stability of NoisyNets output:
$$ D = \frac 1{(p^*+1)N_{a}} (\Sigma_j^{N_{a}}\Sigma_i^{p^*}\sigma_{i,j}^w + \Sigma_j^{N_{a}}\sigma_j^b)  \eqno(8) $$
where \begin{math} p^* \end{math} is the input dimension of the last layer, and \begin{math} N_{a}  \end{math} is the number of output actions. $D$ reflects the noise level of output actions of the agent. It is noticeable that $D$ is differentiable to $\sigma$. And the core idea of noise reduction is that reduce the noise level felicitously to enable the agent perform better, so we can combine the noise level with the TD error to form a new loss function.

Then, we use the sum of the original loss function and $D$ as the new loss function.
$$ L^+(\zeta) =E[E[r_{t}+\gamma \underset{a'}{max}Q^\pi(s_{t+1},a',\varepsilon';\zeta^-)-Q^\pi(s_t,a_t,\varepsilon;\zeta)]+k\cdot{D}]  \eqno(9) $$
where $k$ is a certain proportional coefficient. $k$ is used to control the proportion of update in the TD-error gradient direction and in the attenuation direction. We will introduce the adjustment strategy to control this proportional coefficient in the next section. According to (9), we can further obtain the update direction of parameters in our algorithm.
$$ \bigtriangledown_\zeta L^+(\zeta) =E[\bigtriangledown_\zeta (E[r_{t}+\gamma \underset{a'}{max}Q^\pi(s_{t+1},a',\varepsilon';\zeta^-)-Q^\pi(s_t,a_t,\varepsilon;\zeta)]+k\cdot{D})]  \eqno(10) $$


\begin{figure}[h]
	\begin{spacing}{0.7}	
		\centering
		\subfigure[Learning-independent noise reduction]{
			\begin{minipage}[t]{0.5\linewidth}
				\centering
				\includegraphics[width=2.25in]{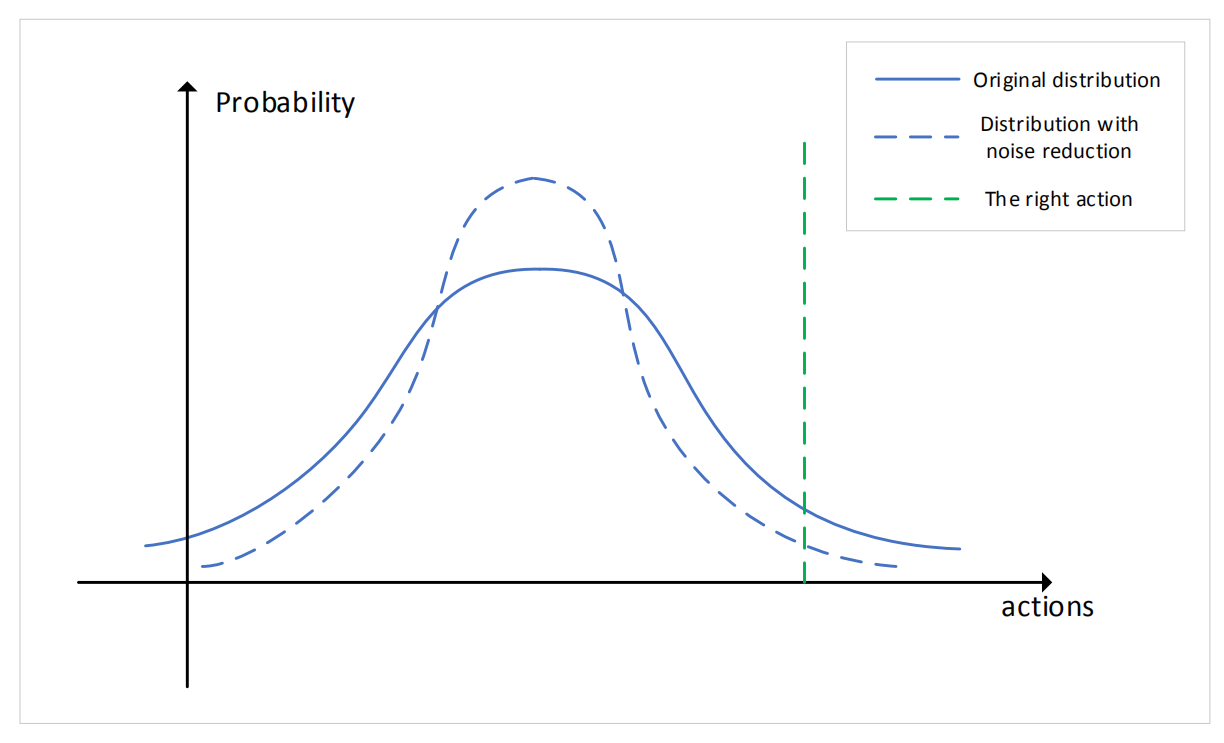}
				\label{fig:noise reduction-a}
			\end{minipage}%
		}%
		\subfigure[Learning-consistent noise reduction]{
			\begin{minipage}[t]{0.5\linewidth}
				\centering
				\includegraphics[width=2.25in]{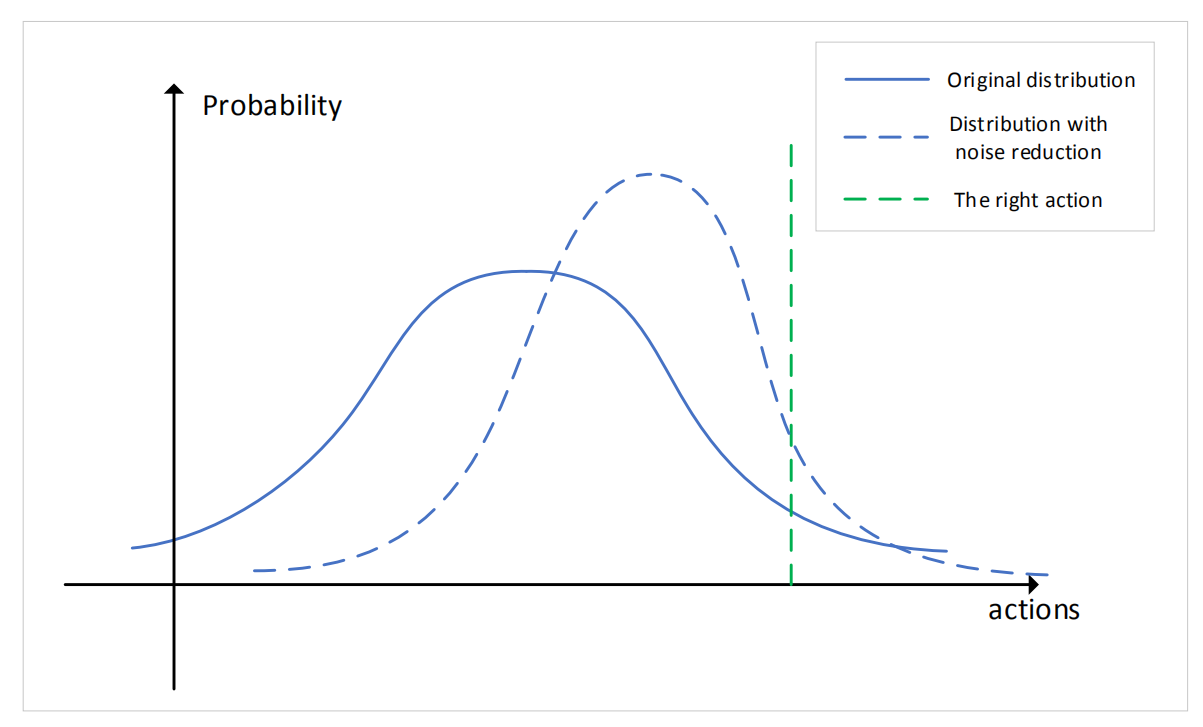}
				\label{fig:noise reduction-b}
			\end{minipage}%
		}%
		\centering
		\caption{Comparison of different noise reduction methods}	
		\label{fig:noise reduction}
	\end{spacing}
\end{figure}

We add the noise reduction mechanism to the loss function of the Q network, which makes the noise reduction process consistent with the learning process. Figure \ref{fig:noise reduction-a} shows the propability of action distribution about a learning-independent noise reduction process. Since the noise reduction process does not occur simultaneously with learning, reducing the output variance alone will decrease the probability that the agent selects the correct action at this time. Our method does not reduce variance independently. The update process of NROWAN-DQN according to equation (10) also suppressed the noise. So the noise reduction process of NROWAN-DQN is more consistent with learning process. In Figure \ref{fig:noise reduction-b}, while the variance of the agent output distribution is reduced, the overall distribution is also closer to the direction of right action.

\subsection{Online weight adjustment}

There are two ways to increase the weight $k$. The first way is to increase $k$ monotonously with the number of training frames. As the number of training frames increases, $k$ finally converages to a certain value.
$$ k = k_{final}-k_{final}\cdot{e^{-N_{f}/a}}  \eqno(11) $$
where \begin{math} k_{final} \end{math} denotes the final expected value of $k$, \begin{math} N_{f} \end{math} denotes the number of frames, and $a$ denotes a factor to control the growth rate of $k$. According to equation (11), when \begin{math} N_{f} \end{math} is very small, $k$ is close to 0. In this case, the agent tends to fully explore, and the update direction of \begin{math} \zeta \end{math} is the TD-error gradient direction. On the contrary, when \begin{math} N_{f} \end{math} is very large, $k$ is close to \begin{math} k_{final} \end{math}. At this time, the agent tends to form a stabilization policy. The \begin{math} \mu \end{math} and \begin{math} \Sigma \end{math} of the inner layer are still updated along the TD-error gradient direction, and the \begin{math} \Sigma \end{math} of the last layer is updated along the sum gradient direction of the TD-error and the $D$ vector.

\begin{algorithm}[!t]
	\begin{spacing}{1.6}	
	\end{spacing}
	
	{\bf Algorithm 1 }{NROWAN-DQN \\} 
	{------------------------------------------------------------------------------------------------------}     
	\begin{spacing}{0.7}	
	\end{spacing}
	\begin{spacing}{1.6}
		\hspace*{0.02in} {\bf Input: }{learning rate $\alpha$, update frequency $N_c$, min-frame to start learning $t_0$, } \\    
		\hspace*{0.02in} memory capacity $N$, budget $T$, final weighting factor $k_{final}$\\
		\hspace*{0.02in} {\bf Output: $Q(s,a,\varepsilon^{'};\zeta^-)$}     
		
		\begin{algorithmic}[1]
			\State Initialize replay memory with capacity $N$     
			\State Initialize online Q-net with $\varepsilon$ and $(\mu, \Sigma)$
			\State Initialize target Q-net with $\varepsilon'\gets\varepsilon $ and  $\zeta^{-}\gets\zeta$
			\State Observe $sup(R), inf(R) \sim Env$ 
			\For{$t$ = 1 to $T$}     
			\State Select $ a_t = arg max_aQ(s_t,a,\varepsilon;\zeta) $ 
			\State Execute $ a_t $ and observe $ (r_t, s_{t+1}) \sim Env $
			\State Store experience $ (s_t,a_t,r_t,s_{t+1}) $ in replay buffer
			\State Calculate $D$ of online Q-net according to equation (8)
			\State $ r_t^+ \gets r_t^++r_t $ and calculate $k$ according to $ r_t^+,sup(R)$ and $inf(R) $
			\If{$ s_{t+1} $ is a terminal state}     
			\State $ r_t^+ \gets 0 $
			\EndIf
			\If{$ t > t_0 $}     
			\State Sample random minibatch of experience from replay buffer
			\State \begin{math}\bigtriangledown \gets \bigtriangledown_\zeta{(r_{t}+\gamma max_{a'}Q^\pi(s_{t+1},a';\theta^-)-Q^\pi(s_t,a_t;\theta)+k\cdot{D})} \end{math} 
			\State $\zeta \gets \zeta + \alpha\cdot\bigtriangledown$ 
			\EndIf
			\If{$ t $  mod  $ N_c  =  0 $}     
			\State $\zeta^{-}\gets\zeta$
			\EndIf
			\EndFor
			
		\end{algorithmic}
	\end{spacing}
\end{algorithm}

The second way to increase the weight $k$ is to adjust it online based on the reward:
$$ k = k_{final}\cdot{(r_t^+-inf(R))/(sup(R)-inf(R))}  \eqno(12) $$
where \begin{math} r_t^+ = \Sigma_{i=0}^{i=t}r_i \end{math} denotes the current reward, \begin{math} sup(R) \end{math} denotes the maximum reward and \begin{math} inf(R) \end{math} denotes the minimum reward, respectively. In the early stage of learning, an agent cannot obtain a high reward in general. \begin{math} r_t^+ \end{math} is usually close to \begin{math} inf(R) \end{math}, so $k$ is close to 0. At this time, the parameters are updated in the TD-error gradient direction. When the learning reaches to a certain stage, the \begin{math} \mu \end{math} and \begin{math} \Sigma \end{math} of the inner layer are updated in the TD-error gradient direction, while the \begin{math} \Sigma \end{math} of the last layer is alternately updated in the TD-error gradient direction and the \begin{math} \bigtriangledown_\sigma{L^+} \end{math} direction. This is because at the beginning of each round of a game, $r$ is close to \begin{math} inf(R) \end{math}, $k$ is close to 0, and \begin{math} \Sigma \end{math} is updated in the TD-error gradient direction. With a round of the game going on, the current reward \begin{math} r_t^+ \end{math} increases, the weight $k$ gets closer and closer to \begin{math} k_{final} \end{math}, and the \begin{math} \Sigma \end{math} of the last layer is updated in the \begin{math} \bigtriangledown_\sigma{L^+} \end{math} direction. After the current round ends up, the next round begins. At this point, \begin{math} r_t^+ \end{math} becomes a small value again, and a new loop starts. Our experiments show that the training with this updating method can produce a more stable policy.

The selection of \begin{math} inf(R) \end{math} and \begin{math} sup(R) \end{math} does not need to be too strict. In some virtual environments, \begin{math} inf(R) \end{math} and \begin{math} sup(R) \end{math} can be easily obtained. When the environment rewards are not clear, \begin{math} inf(R) \end{math} can be set to a reward that a randomly initialized policy can achieve, and \begin{math} sup(R) \end{math} can be set to the highest reward that the current algorithm can achieve in this environment. It is worth noting that the closer \begin{math} inf(R) \end{math} and \begin{math} sup(R) \end{math} are, the more carefully the parameter \begin{math} k_{final} \end{math} needs to be adjusted. In the experimental part, we will only decay the noise with online weight adjustment.

Algorithm 1 describes the training process of NROWAN-DQN. This algorithm follows the original framework of the NoisyNet-DQN algorithm \cite{Ref20173}. In Algorithm 1, since the $k$ is adjusted online based on the reward,  additional input \begin{math} k_{final} \end{math}  is required. It is also necessary to obtain the range of environmental rewards. When calculating the current reward in line 10, the $k$ is also calculated according to equation (12). In line 15, the updating direction of parameters is calculated. And in line 16, parameters are updated according to the updating direction and the learning rate.

\section{Experiments}

This section provides a description of the experimental environment, parameters setting, results of comparative experiments, and some analytical results.

\subsection{Environments}

\begin{figure}[h]
	\begin{spacing}{0.7}	
		\centering
		\subfigure[Cartpole]{
			\begin{minipage}[t]{0.5\linewidth}
				\centering
				\includegraphics[width=2in]{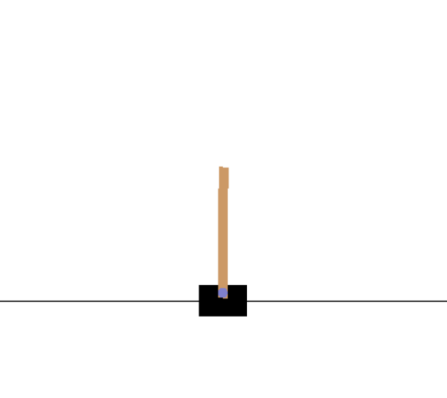}
			\end{minipage}%
		}%
		\subfigure[Pong]{
			\begin{minipage}[t]{0.5\linewidth}
				\centering
				\includegraphics[width=2in]{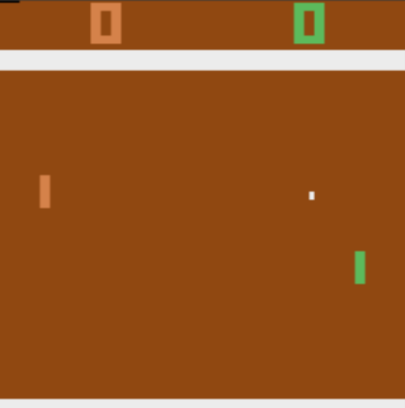}
			\end{minipage}%
		}%
		
		\subfigure[MountainCar]{
			\begin{minipage}[t]{0.5\linewidth}
				\centering
				\includegraphics[width=2in]{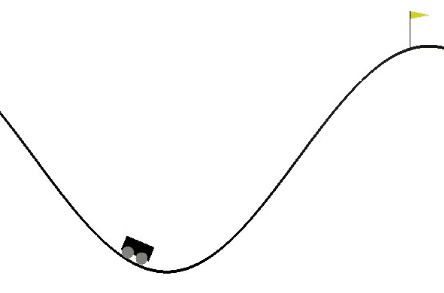}
			\end{minipage}%
		}%
		\subfigure[Acrobot]{
			\begin{minipage}[t]{0.5\linewidth}
				\centering
				\includegraphics[width=2in]{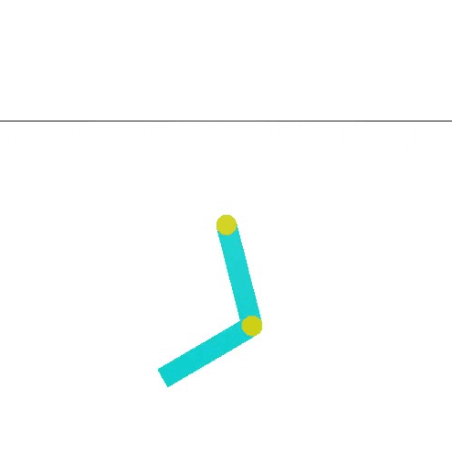}
			\end{minipage}%
		}%
		\centering
		\caption{Four typical reinforcement learning environments}	
		\label{fig:4game}
	\end{spacing}
	
\end{figure}

Our experiments were conducted in OpenAI Gym \cite{Ref20164}, which provides a collection of environment for reinforcement learning tasks. We selected four typical environments for testing. These four environments are Cartpole, Pong, MountainCar and Acrobot. \footnote{The codes involved in this section can be found on the website https://github.com/HCodeRunner/noisy-networks-with-deterministic-factor.}

$\bullet$ In Cartpole, the environment provides an observation state to an agent at each frame. The observation state is a 4-tuple, which consists of the position of a cart, the speed of the cart, the angle of a pole, and the angular velocity of a pole. The agent can choose to move the cart to the left or the right. After the agent selects an action, the environment returns a ``+1" reward. When the total reward reaches ``+200" or the pole is more than 15 degrees from vertical, the game is over.  \footnote{Descriptions of these environments can be found on website: http://gym.openai.com/envs/CartPole-v1/}

$\bullet$ Pong provides an agent with an end-to-end learning environment. In each frame, the environment delivers an RGB image to the agent. The agent controls a racket to hit a ball. If the opponent misses the ball, the agent obtains a ``+1" reward. If the agent misses the ball, the agent obtains a ``-1" reward. When the total reward reaches ``+21" or ``-21", the game is over.  \footnote{Descriptions of these environments can be found on website: http://gym.openai.com/envs/Pong-v0/}

$\bullet$ In MountainCar, a car is positioned between two ``mountains". The goal is to drive the car up to the mountain on the right; however, the car's engine is not strong enough to scale the mountain in a single pass. Therefore, the only way to succeed is to drive back and forth to build up momentum. \footnote{Description of this environment is cited from: http://gym.openai.com/envs/MountainCarContinuous-v0/}
	
$\bullet$ The Acrobot system includes two joints and two links, where the joint between the two links is actuated. Initially, the links are hanging downwards, and the goal is to swing the end of the lower link up to a given height. \footnote{Description of this environment is cited from: http://gym.openai.com/envs/Acrobot-v1/}

Some screenshots of these games are shown in Figure \ref{fig:4game}.

\subsection{Hyper-parameters}

\begin{table}[!htbp]
	\begin{spacing}{1.4}
		\caption{Pre-processing, network structure, and public hyper-parameters setting}  
		\begin{spacing}{1.6}	
		\end{spacing}
		\centering
		\resizebox{\textwidth}{!}{
			\begin{tabular}{llll}
				\toprule
				{Hyper-parameters}& Pong& Others(Cartpole, etc.) & Description\\
				\hline
				\multirow{2}{25mm}{Grey-scaling}& \multirow{2}{20mm}{True}& \multirow{2}{20mm}{-----}& \multirow{2}{60mm}{Whether the observations are converted into single-channel images}\\
				\\
				\hline
				\multirow{2}{25mm}{Observation down-sampling}& \multirow{2}{20mm}{(84,84)}& \multirow{2}{20mm}{-----}& \multirow{2}{60mm}{Adjust the size of the observation}\\
				\\
				\hline
				\multirow{2}{25mm}{Frames stack}& \multirow{2}{20mm}{1}& \multirow{2}{20mm}{1}& \multirow{2}{60mm}{How many frames are stacked as one input frame}\\
				\\
				\hline
				\multirow{2}{25mm}{Action repetitions}& \multirow{2}{20mm}{1}& \multirow{2}{20mm}{1}& \multirow{2}{60mm}{How many times an action is repeated}\\
				\\
				\hline
				\multirow{2}{25mm}{Reward clipping}& \multirow{2}{20mm}{False}& \multirow{2}{20mm}{False}& \multirow{2}{60mm}{Whether rewards is clipped or not}\\
				\\
				\hline
				\multirow{2}{25mm}{Q network: channels}& \multirow{2}{20mm}{32,64,64}& \multirow{2}{20mm}{-----}& \multirow{2}{60mm}{Number of filters for each layer}\\
				\\
				\hline
				\multirow{2}{25mm}{Q network: filter size}& \multirow{2}{20mm}{8$\times$8,4$\times$4,3$\times$3}& \multirow{2}{20mm}{-----}& \multirow{2}{60mm}{Size of the filter for each layer}\\
				\\
				\hline
				\multirow{2}{26mm}{Q network: stride}& \multirow{2}{20mm}{4,2,1}& \multirow{2}{20mm}{-----}& \multirow{2}{60mm}{Step size of the filter in each layer}\\\\
				\hline
				\multirow{2}{26mm}{Q network: hidden layer}& \multirow{2}{20mm}{512,512}& \multirow{2}{20mm}{128,128}& \multirow{2}{60mm}{Number of neurons in each hidden layer}\\\\
				\hline
				\multirow{2}{26mm}{Q network: output layer}& \multirow{2}{20mm}{Numbers of actions}& \multirow{2}{20mm}{Numbers of actions}& \multirow{2}{60mm}{Number of neurons in the output layer}\\\\
				\hline
				\multirow{2}{26mm}{Activation function}& \multirow{2}{20mm}{ReLu}& \multirow{2}{20mm}{ReLu}& \multirow{2}{60mm}{Activation function in each layer except the output layer}\\\\
				\hline
				\multirow{2}{26mm}{Budget}& \multirow{2}{20mm}{1,000,000}& \multirow{2}{20mm}{30,000}& \multirow{2}{60mm}{Number of environment frames for training}\\\\
				\hline
				\multirow{2}{26mm}{Batch size}& \multirow{2}{20mm}{32}& \multirow{2}{20mm}{32}& \multirow{2}{60mm}{Number of samples used for gradient descent}\\\\
				\hline
				\multirow{2}{26mm}{Gamma($\gamma$)}& \multirow{2}{20mm}{0.99}& \multirow{2}{20mm}{0.99}& \multirow{2}{60mm}{Reward discount factor}\\\\
				\hline
				\multirow{2}{26mm}{Update frequency}& \multirow{2}{20mm}{every 1000 step}& \multirow{2}{20mm}{every 1000 step}& \multirow{2}{60mm}{The frequency that the target network is updated to the current network}\\\\
				\hline
				\multirow{2}{26mm}{Min-frame to start learning}& \multirow{2}{20mm}{10000}& \multirow{2}{20mm}{32}& \multirow{2}{60mm}{Number of frames that start learning}\\\\
				\hline
				\multirow{2}{26mm}{Memory capacity}& \multirow{2}{20mm}{100,000}& \multirow{2}{20mm}{10,000}& \multirow{2}{60mm}{Size of the replay buffer}\\\\
				\hline
				\multirow{2}{26mm}{Frequency of learning}& \multirow{2}{20mm}{every 1 step}& \multirow{2}{20mm}{every 1 step}& \multirow{2}{60mm}{Frequency of sampling and performing gradient updates}\\\\
				\bottomrule
		\end{tabular}  	}
	\end{spacing}
\end{table}

In our experiments, hyper-parameters are designed based on Hessel's, Mnih's, and Fortunato's algorithms \cite{Ref20181} \cite{Ref20151} \cite{Ref20173}, and the network structure is exactly the same as that in Hessel's algorithm. We have fine-tuned some of the hyper-parameters in order to fit our experimental situation better. For example, in Cartpole and Pong, it takes 30K frames and 1M frames respectively to run a round of experiments. So it is inappropriate to use a replay buffer with 1M capacity as the setting in Mnih's algorithm \cite{Ref20151}. When we conducted a comparative experiment, hyper-parameters of all the public parts of the algorithms are consistent in each comparison. In order to ensure the reproducibility of the experiment, we attach a detailed report in Table 1, which includes pre-processing setting, network structure setting, and public hyper-parameters setting.

In the original environment, each frame is a 3-channel color image. These pictures will be converted into single-channel grayscale images. Then these pictures will be scaled to (84, 84) and transmitted to the agent. Since the scale of the four problems is small, we don't stack frames. During the learning process, each frame will be treated as a separate state for the agent. Correspondingly, the action will not be repeated. Since there are not involving statistics among different games, we do not clip the rewards. Since Cartpole, MountainCar and Acrobot are not involving high-dimensional state input, the original information from these environments will be transmitted to the agent without any processing.

Our Q network structure is the same as the common Q network structure in the reinforcement learning. There are three convolution layers in our Q network. The first layer has 32 channels, the convolution kernel size is 8$ \times $8, and the step size is 4. The second layer has 64 channels, the convolution kernel size is 4$ \times $4, and the step size is 2. The third layer has 64 channels, the convolution kernel size is 3$ \times $3, and the size is 1. After convolution processing, the pixel matrix is changed into a one-dimensional vector and delivered to the fully connected layer. The fully connected layer contains a hidden layer and an output layer. The hidden layer has 512 neurous. The number of neurous in the output layer is the same with the number of environmental actions. Except the output layer, all the layers take the ReLu function as their activation function.

Since Cartpole, MountainCar and Acrobot do not need to process high-dimensional information, the state information provided by the Gym can be directly delivered to the fully connected layer. The fully connected layer has two hidden layers and an output layer. Each hidden layer has 128 neurons. The output layer has the same number of neurons as the number of environmental actions. Similarly, except the output layer, all the layers take the ReLu function as their activation function.

1M frames are required in each round of learning in Pong, and the replay buffer size is 100K. In the first 10K frames, gradient update process of Q network is not performed. The agent take actions using a random initialization network. During this process, the 4-tuple \begin{math} (s_t,a_t,r_{t},s_{t+1}) \end{math} is pushed into the replay buffer at each time step. After 10K frames, agent randomly samples 32 tuples from the replay buffer at each time step, and performs batch random gradient descent. Every 1K frames, the parameters of the target network are updated to the parameters of the current online network.

For Cartpole, MountainCar and Acrobot, each round of learning only takes 30K frames, and the replay buffer size is 10K. When the number of quaternions in the replay buffer is enough for the agent to sample, the learning starts. Agent randomly samples 32 tuples from the replay buffer at each time step, and performs batch random gradient descent on the parameters in the online network. The parameters of the target network are updated to the parameters of the current online network every 1K frames.

In addition to the above setting, other hyper-parameters are involved in our algorithm. When using the online adjustment strategy, \begin{math} k_{final} \end{math} is set to 4.0. In the following experiments, \begin{math} k_{final} \end{math} will be set to a value in the range of \{2.0, 2.5, 3.0, 3.5, 4.0, 4.5, 5.0, 5.5, 6.0\}.

Agents using noisy networks also have hyper-parameters setting that is not available in the original DQN. In our experiment, the initial value \begin{math} \sigma_0 \end{math} is set to 0.4. The noise factor \begin{math} \varepsilon_i \end{math} and \begin{math} \varepsilon_j \end{math} are randomly generated according to a normal distribution. The shape of the normal distribution is determined by the dimension of input or output. All the agents use the Adam optimizer \cite{Ref20141} to update the network parameters. The learning rate \begin{math} \alpha \end{math} is set to 0.0001. Table 2 shows the learning and noise parameters setting. In the following experiments, \begin{math} \alpha \end{math} will be set to a value in the range of \{0.0001, 0.000075, 0.000025\}.

\begin{table}[htb]
	\begin{spacing}{1.6}
		\caption{Learning and noise parameters setting}  
		\begin{spacing}{1.6}	
		\end{spacing}
		\centering
		\begin{tabular}{ccccc}
			\hline
			{Hyper-parameters}& Cartpole& Pong& MountainCar& Acrobot\\
			\hline
			{Learning rate}& {0.0001}& {0.0001}& {0.001}& {0.001}\\
			{Adam $\epsilon$}& {0}& {0}& {0}& {0}\\
			{Adam $\beta$}& {(0.9, 0.999)}& {(0.9, 0.999)}& {(0.9, 0.999)}& {(0.9, 0.999)}\\
			{Noisy linear $\sigma_0$}& {0.4}& {0.4}& {0.4}& {0.4}\\
			{$k_{final}$}& {4.0}& {4.0}& {4.0}& {4.0}\\
			\hline
		\end{tabular}   	
	\end{spacing}
\end{table}

\subsection{Score comparison}

Table 3 shows the performance of three algorithms (i.e., DQN, NoisyNet-DQN, and NROWAN-DQN) in Cartpole, Pong, MountainCar and Acrobot. With the same parameters, we trained five instances for each algorithm, and each instance ran 64 rounds. We calculated the average score of these 64 rounds as the score of the instance. Then we calculated the average score of these five instance as the final score. The standard deviation of the scores were calculated in a similar way. NoisyNet-DQN reported in Table 3 uses online noise adjustment to decay the noise. As shown in Table 3, NROWAN-DQN scored an average of 187.04 in Cartpole, which is 16.56 higher than DQN, and 22.08 higher than NoisyNet-DQN, and the average standard deviation is nearly 3 times smaller than DQN and NoisyNet-DQN. In Pong, NROWAN-DQN has an average score of 18.80, which is 1.74 higher than DQN, 0.86 higher than NoisyNet-DQN, and the average standard deviation is 0.5 less than DQN, 0.2 less than NoisyNet-DQN. In MountainCar, NROWAN-DQN has an average score of -121.85, which is 10.05 higher than DQN, 6.52 higher than NoisyNet-DQN, and the average standard deviation is 1.21 less than DQN, 2.09 less than NoisyNet-DQN. In Acrobot, NROWAN-DQN has an average score of -84.41, which is 2.83 higher than DQN, 2.16 higher than NoisyNet-DQN, and the average standard deviation is 6.57 less than DQN, 13.74 less than NoisyNet-DQN. The results of these experiments show that NROWAN-DQN has more stable and higher performance.

In Table 3, NoisyNet-DQN performs better than DQN in Pong, which is consistent with the experimental results of Fortunato et al. \cite{Ref20151}. However, it is unexpected that NoisyNet-DQN scores less than DQN in Cartpole. After analyzing the experimental environment, we found it is because this environment is more sensitive to noise actions. In Cartpole, for example, when the pole tilts to left, moving the cart to the left due to noise is likely to cause the game to be lost. Other environment is less sensitive to noise action. For example, in Pong, when the racket can hit the ball in the original position, it is likely that agents will still be able to receive the ball when agents move it a little bit due to noise. What's more, the large state-action space in Pong needs to be more fully explored for the environment. With better exploration mechanism, NoisyNet-DQN outperforms DQN. Because of NROWAN-DQN exploring more fully and mitigating the effects of noise, it outperforms the other two algorithms in all environments.

\begin{table}  
	\begin{spacing}{1.6}	
		\caption{Performance comparison}  
		\begin{spacing}{1.6}	
		\end{spacing}
		\centering
		\begin{tabular*}{11cm}{cccc}
			\hline  
			{Problem}& {DQN}& {NoisyNet-DQN}& {NROWAN-DQN}\\
			\hline
			{Cartpole}& {170.49$ \pm $35.86}& {164.96$ \pm $31.56}& \textbf{187.04$ \pm $13.99}\\
			{Pong}& {17.07$ \pm $3.36}& {17.95$ \pm $3.08}& \textbf{18.81$ \pm $2.87}\\
			{MountainCar}& {-131.90$ \pm $21.09}& {-128.37$ \pm $21.97}& \textbf{-121.85$ \pm $19.88}\\
			{Acrobot}& {-87.24$ \pm $22.33}& {-86.57$ \pm $29.32}& \textbf{-84.41$ \pm $15.58}\\
			\hline
		\end{tabular*}  	
	\end{spacing}
\end{table}

\subsection{Score change in learning}

Figure 6 shows the average reward trend of NoisyNet-DQN and NROWAN-DQN in Cartpole, Pong, MountainCar and Acrobot. In Pong and MountainCar, NROWAN-DQN performs better than NoisyNet-DQN during the learning process. This is because the deterministic factor makes it easier for agents to form stable policies when learning. In Cartpole, NROWAN-DQN can effectively learn in a noise sensitive environment, so its learning curve can converge to higher values. Since the Acrobot environment is simpler, both NROWAN-DQN and NoisyNet-DQN can learn quickly in this environment, and the learning curves of both can converge to approximately the same value, so it is difficult to compare NROWAN-DQN and NoisyNet-DQN in this environment. But overall, we can conclude that the learning ability of NROWAN-DQN is better than NoisyNet-DQN.

\begin{figure}[h]
	\begin{spacing}{0.7}	
		\centering
		\subfigure[Cartpole]{
			\begin{minipage}[t]{0.5\linewidth}
				\centering
				\includegraphics[width=2in]{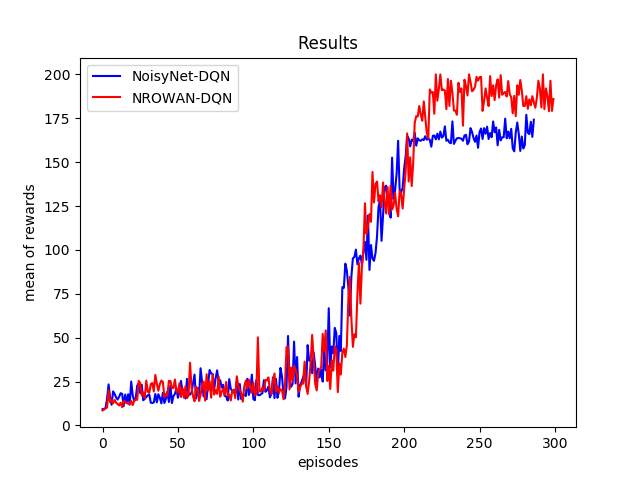}
			\end{minipage}%
		}%
		\subfigure[Pong]{
			\begin{minipage}[t]{0.5\linewidth}
				\centering
				\includegraphics[width=2in]{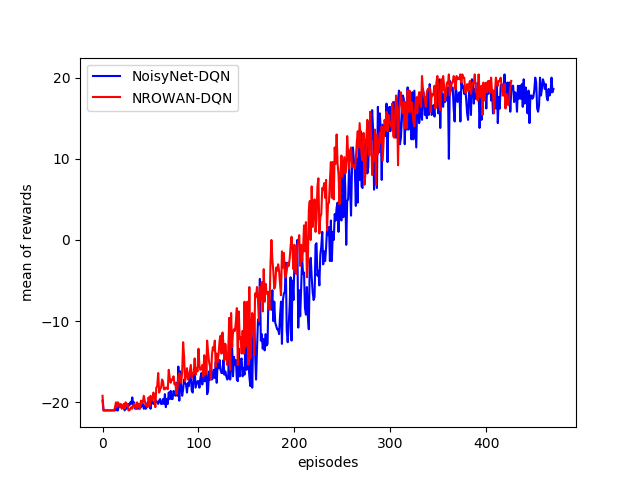}
			\end{minipage}%
		}%

		\subfigure[MountainCar]{
			\begin{minipage}[t]{0.5\linewidth}
				\centering
				\includegraphics[width=2in]{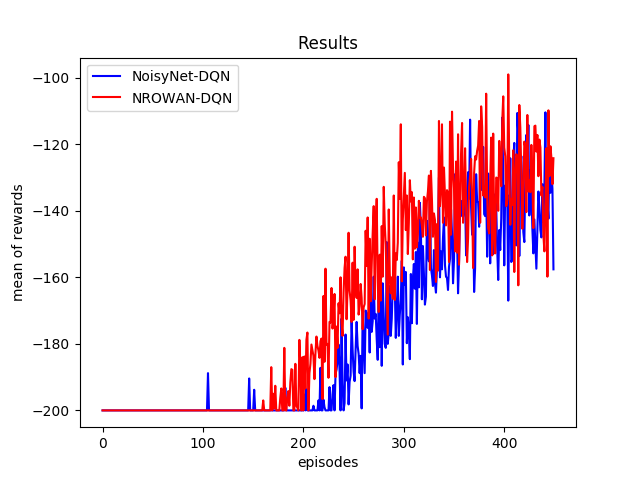}
			\end{minipage}%
		}%
		\subfigure[Acrobot]{
			\begin{minipage}[t]{0.5\linewidth}
				\centering
				\includegraphics[width=2in]{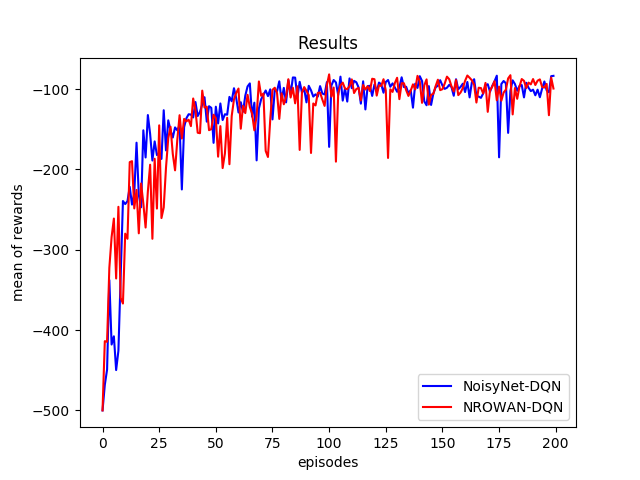}
			\end{minipage}%
		}%
		\centering
		\caption{Learning curves of NoisyNet-DQN and NROWAN-DQN in Cartpole, Pong, MountainCar and Acrobot}	
	\end{spacing}
\end{figure}

\subsection{Final $k$ value and learning rate}

\begin{figure}[!htbp]
	\centering  
	\includegraphics[width=0.9\linewidth]{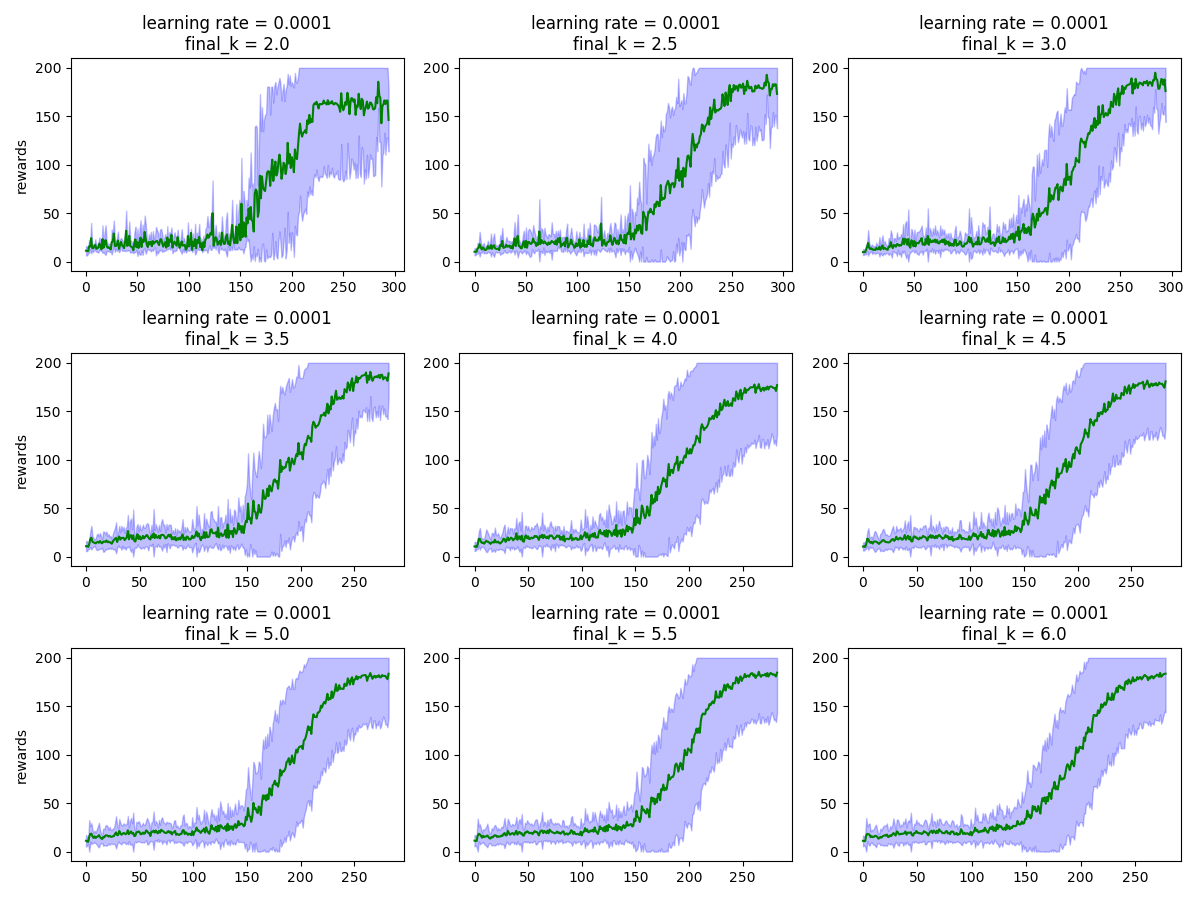}  
	\caption{Score results with different $k_{final}$ in Cartpole when learning rate is 0.0001}  
	\label{fig:rl100}   
\end{figure}

\begin{figure}[!htbp]
	\centering  
	\includegraphics[width=0.9\linewidth]{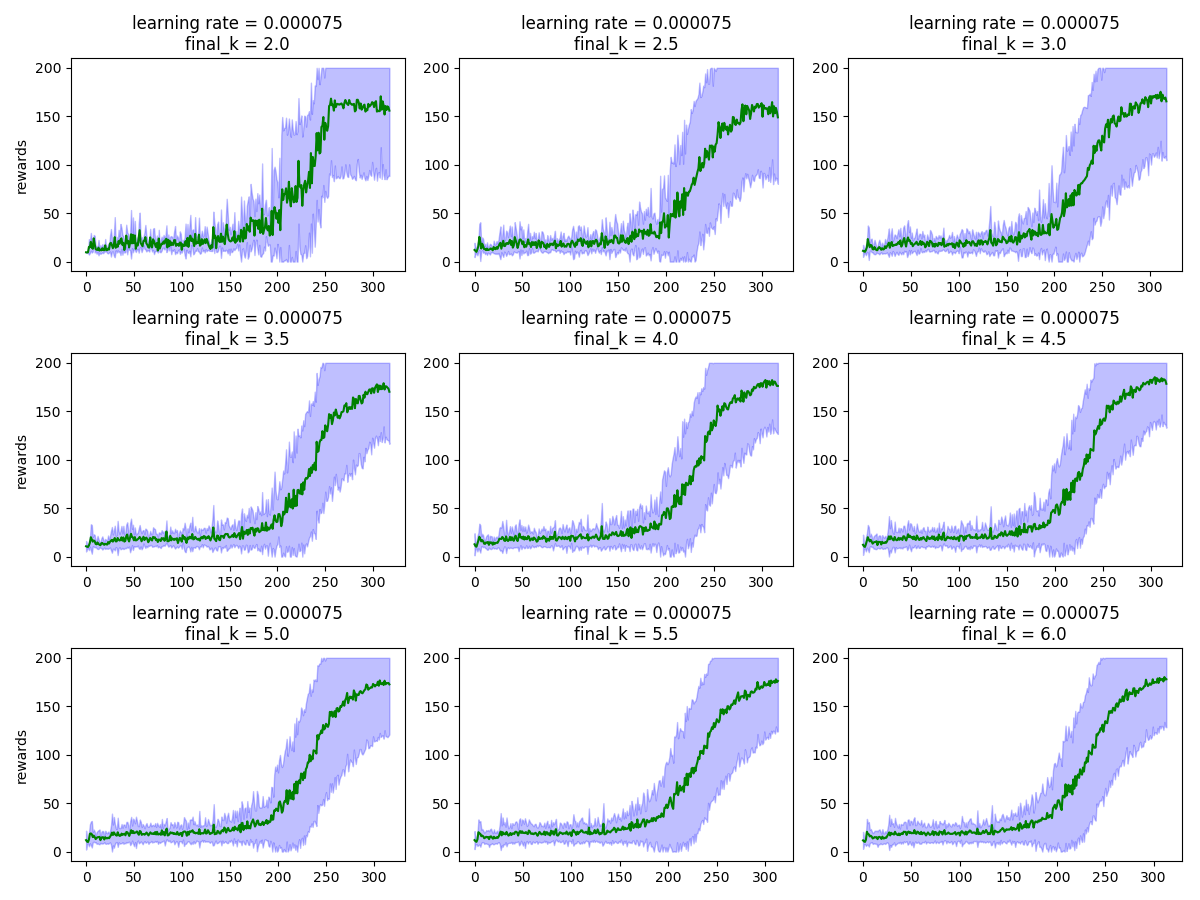}  
	\caption{Score results with different $k_{final}$ in Cartpole when learning rate is 0.000075}  
	\label{fig:rl075}   
\end{figure}

\begin{figure}[!htbp]
	\centering  
	\includegraphics[width=1.0\linewidth]{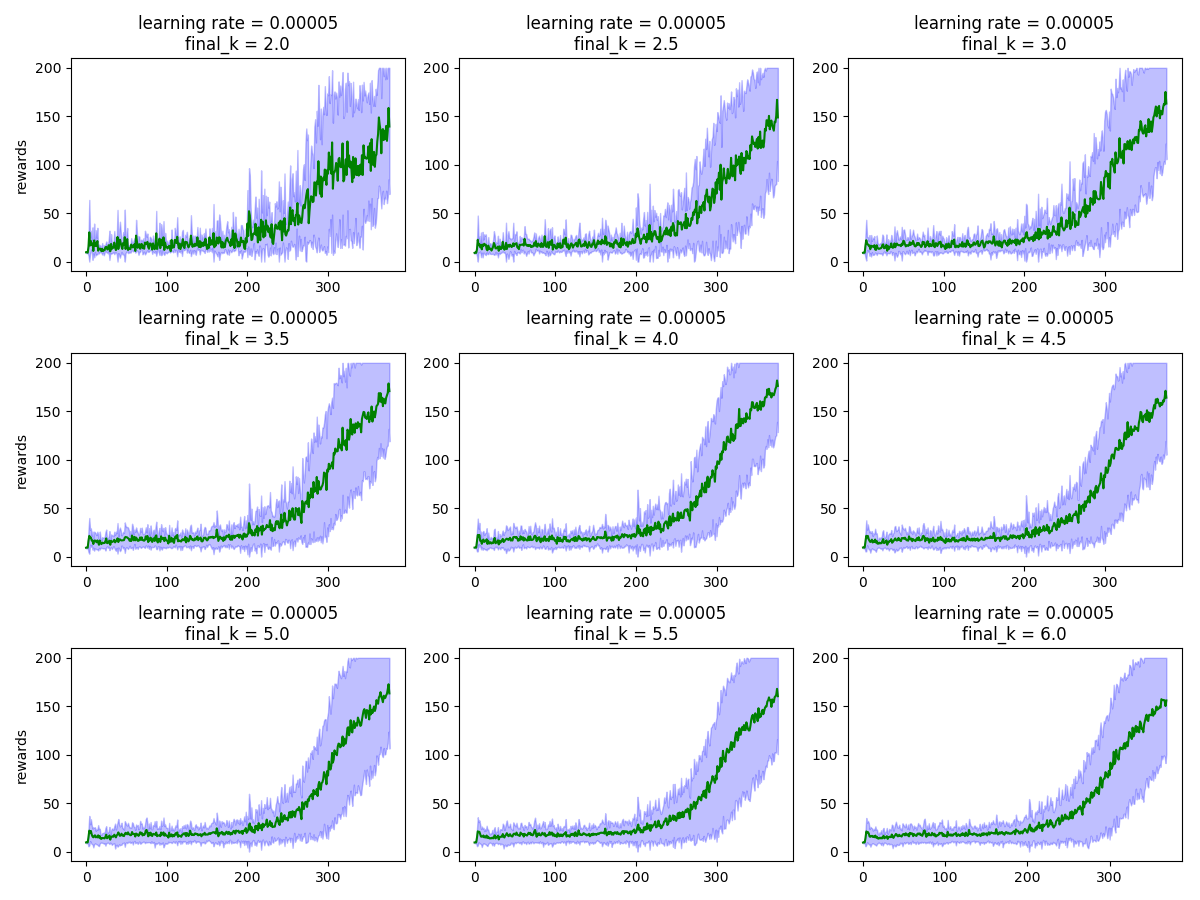}  
	\caption{Score results with different $k_{final}$ in Cartpole when learning rate is 0.00005}  
	\label{fig:rl050}   
\end{figure}

\begin{figure}[!htbp]
	\centering  
	\includegraphics[width=0.7\linewidth]{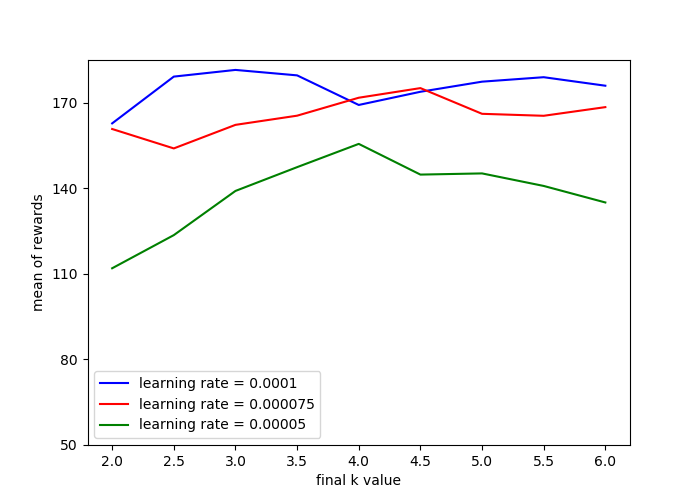}  
	\caption{Score results with different learning rates and different $k_{final}$ values in Cartpole}  
	\label{fig:3to1}   
\end{figure}

Our algorithm introduces a new hyper-parameter \begin{math} k_{final} \end{math}, which indicates the final value of $k$ in the learning process. We analyzed the relationship between \begin{math} k_{final} \end{math} and learning rate in the low-dimensional Cartpole environment, and tested the appropriate value of \begin{math} k_{final} \end{math} in the high-dimensional Pong environment.

Because \begin{math} k_{final} \end{math} and learning rate jointly control the  parameter update step size in the direction of generating a stable policy,  \begin{math} k_{final} \end{math} should be adjusted along with the learning rate. Figures \ref{fig:rl100} - \ref{fig:rl050} show the learning curve of the agent in Cartpole with three different learning rates and nine different \begin{math} k_{final} \end{math} values. As we can see, too small \begin{math} k_{final} \end{math} will lead to instability, and the final results have a large variance; too large $k_{final}$ will cause output actions of the agent to be too stable, resulting in a decline in the exploration ability, thus making the agent learn slower. Figure \ref{fig:3to1} shows score results with different learning rates and different $k_{final}$ values in Cartpole. The product of $k$ and the learning rate controls the update step size of \begin{math} \bigtriangledown_\sigma{D} \end{math} in equation (10) during learning. More specifically, the learning rate controls the update step size, and $k$ controls the proportion of updates in the TD-error gradient direction and in the \begin{math} \bigtriangledown_\sigma{D} \end{math} direction. When \begin{math} k_{final} \end{math} and the learning rate are both small, the learning efficiency of the agent is low. When \begin{math} k_{final} \end{math} is large and the learning rate is small, the parameters of the agent is over-updated in the direction along which the stable action is generated. After repeated testing, we conclude that the large learning rate is not sensitive to the change of \begin{math} k_{final} \end{math}. When \begin{math} k_{final} \end{math} is 4, the algorithm is less sensitive to the change of learning rate. Therefore, it is recommended that \begin{math} k_{final} \end{math} be set to 4.

\begin{figure}[!htbp]
	\centering  
	\includegraphics[width=1.0\linewidth]{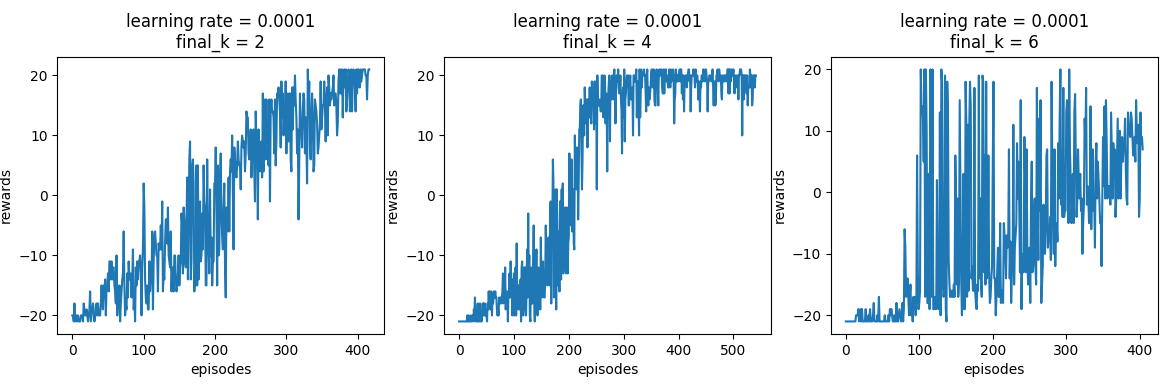}  
	\caption{Learning curves for different final $k$ values in Pong}  
	\label{fig:3pong}   
\end{figure}

We also tested the effect of different \begin{math} k_{final} \end{math} values on the learning outcome in a high dimensional environment. Figure \ref{fig:3pong} show that when $k_{final}$ is set to 4, the agent still has good learning ability. However, because Pong is less sensitive to noise than Cartpole, too small $k_{final}$ will not cause the agent in Pong to be unstable. On the contrary, because $k_{final}$ is small, the agent cannot form a stable policy in time, as a result, the learning speed of the agent is reduced. Too large $k_{final}$ causes the parameter to be updated  too much in the direction of noise reduction, which destroys the update process of the agent along the learning direction, resulting in unstable performance of the agent. The magnitude of $k_{final}$ not only controls the balance between exploration and utilization of the agent, but also controls the proportion of parameters update in the direction of noise reduction and learning direction. So $k_{final}$ needs to be carefully set.

\section{Discussion and conclusion}

We have demonstrated that adding the deterministic factor to the loss function of noisy networks can enable the agent get a higher score and better stability. Moreover, we found that NROWAN-DQN outperforms NoisyNet-DQN and DQN in a noise-sensitive environment. Due to instability, DQN cannot be used in some dangerous situations which requires high action precision. In such situations, NoisyNet-DQN is also not appropriate, since it adopts unstable behavior policy. Compared with NoisyNet-DQN, NROWAN-DQN is effective for such situations. Finally, we explore the effect of learning rate and $k_{final}$ on the learning process, and give the recommended value of $k_{final}$.

When developing NoisyNet-DQN, Fortunato et al. also applied noisy networks to DDQN, Dueling and A3C algorithms, which has better performance \cite{Ref20173}. As a component, noisy networks can be used with other components based on Q-value, such as prioritized experience replay \cite{Ref20153} and distributional Q-learning \cite{Ref20176}. What's more, noisy networks have great potential when combined with reinforcement learning algorithms of gradient descent, such as DDPG \cite{Ref20154} and TRPO \cite{Ref20155}. It is also very interesting to study effects and properties when combining our improved NoisyNet with the above-mentioned algorithms or components.

\section*{Acknowledgement}

This work was supported by the National Key R\&D Program of China under Grant No. 2017YFB1003103; the National Natural Science Foundation of China under Grant Nos. 61300049, 61763003; and the Natural Science Research Foundation of Jilin Province of China under Grant Nos. 20180101053JC, 20190201193JC.

{}

\begin{thebibliography}{1}
\expandafter\ifx\csname url\endcsname\relax
  \def\url#1{\texttt{#1}}\fi
\expandafter\ifx\csname urlprefix\endcsname\relax\def\urlprefix{URL }\fi
\expandafter\ifx\csname href\endcsname\relax
  \def\href#1#2{#2} \def\path#1{#1}\fi

\bibitem{Feynman1963118}
R.~Feynman, F.~{Vernon Jr.}, The theory of a general quantum system interacting
  with a linear dissipative system, Annals of Physics 24 (1963) 118--173.
\newblock \href {http://dx.doi.org/10.1016/0003-4916(63)90068-X}
  {\path{doi:10.1016/0003-4916(63)90068-X}}.

\bibitem{Dirac1953888}
P.~Dirac, The lorentz transformation and absolute time, Physica 19~(1-–12)
  (1953) 888--896.
\newblock \href {http://dx.doi.org/10.1016/S0031-8914(53)80099-6}
  {\path{doi:10.1016/S0031-8914(53)80099-6}}.

\end{thebibliography}


\section*{References}
\begin{thebibliography}{}
\bibitem{Ref20166}
Yevgen Chebotar, Karol Hausman, Zhe Su, Gaurav S. Sukhatme, \& Stefan Schaal, Self-supervised regrasping using spatio-temporal tactile features and reinforcement learning, IEEE/RSJ International Conference on Intelligent Robots and Systems (IROS), (2016)

\bibitem{Ref20178}
Marcin Andrychowicz, Filip Wolski, Alex Ray, Jonas Schneider, Rachel Fong, Peter Welinder, Bob McGrew, Josh Tobin, OpenAI Pieter Abbeel, \& Wojciech Zaremba, Hindsight experience replay, Advances in Neural Information Processing Systems, (pp. 5048-5058) (2017)

\bibitem{Ref20179}
Katyal, K., Wang, I., \& Burlina, P., Leveraging deep reinforcement learning for reaching robotic tasks, Proceedings of the IEEE Conference on Computer Vision and Pattern Recognition Workshops, (pp. 18-19) (2017)

\bibitem{Ref20167}
David Silver, Aja Huang, Chris J. Maddison, Arthur Guez, Laurent Sifre, George van den Driessche, Julian Schrittwieser, Ioannis Antonoglou, Veda Panneershelvam, Marc Lanctot, Sander Dieleman, Dominik Grewe, John Nham, Nal Kalchbrenner, Ilya Sutskever, Timothy Lillicrap, Madeleine Leach, Koray Kavukcuoglu, Thore Graepel \& Demis Hassabis, Mastering the game of Go with deep neural networks and tree search, Nature, 529(7587): 484. (2016)

\bibitem{Ref201710}
Lewis, M., Yarats, D., Dauphin, Y., Parikh, D., \& Batra, D. Deal or no deal? End-to-end learning of negotiation dialogues. In Proceedings of the 2017 Conference on Empirical Methods in Natural Language Processing (pp. 2443-2453) (2017).

\bibitem{Ref20183}
Gellert Weisz, 	Pawel Budzianowski, Pei-Hao Su, \& Milica Gasic, 
Sample efficient deep reinforcement learning for dialogue systems with large action spaces, IEEE/ACM Transactions on Audio, Speech and Language Processing (TASLP), 26(11): 2083-2097. (2018)

\bibitem{Ref20156}
V. Derhami, J. Paksima, \& H. Khajah, Web pages ranking algorithm based on reinforcement learning and user feedback, Journal of AI and Data Mining, (pp. 157-168) (2015)

\bibitem{Ref20101}
Jaksch T, Ortner R, \& Auer P, Near-optimal regret bounds for reinforcement learning, Journal of Machine Learning Research, (pp. 1563-1600) (2010)

\bibitem{Ref20157}
Dann C, \& Brunskill E, Sample complexity of episodic fixed-horizon reinforcement learning, Advances in Neural Information Processing Systems, (pp. 2818-2826) (2015)

\bibitem{Ref20081}
Strehl A L, \& Littman M L, An analysis of model-based interval estimation for Markov decision processes, Journal of Computer and System Sciences, (pp. 1309-1331) (2008)

\bibitem{Ref201711}
Georg Ostrovski, Marc G. Bellemare, Aäron van den Oord, \& Rémi Munos, Count-based exploration with neural density models, Proceedings of the 34th International Conference on Machine Learning, (pp. 2721-2730) (2017)

\bibitem{Ref201712}
Justin Fu, John Co-Reyes, \& Sergey Levine, EX2: Exploration with exemplar models for deep reinforcement learning, Advances in Neural Information Processing Systems, (pp. 2577-2587) (2017)

\bibitem{Ref20191}
Rui Zhao, \& Volker Tresp, Curiosity-driven experience prioritization via density estimation, arXiv preprint arXiv:1902.08039 (2019)

\bibitem{Ref20111}
Sun Y, Gomez F, \& Schmidhuber J, Planning to be surprised: Optimal bayesian exploration in dynamic environments, International Conference on Artificial General Intelligence (2011)

\bibitem{Ref20102}
Schmidhuber, \& J. (2010). Formal theory of creativity, fun, and intrinsic motivation (1990-2010). IEEE Transactions on Autonomous Mental Development, 2(3), 230-247.

\bibitem{Ref201713}
Deepak Pathak, Pulkit Agrawal, Alexei A. Efros, \& Trevor Darrell, Curiosity-driven exploration by self-supervised prediction, The IEEE Conference on Computer Vision and Pattern Recognition (CVPR) Workshops, (pp. 16-17) (2017)

\bibitem{Ref201714}
Xiong, W., Hoang, T., \& Wang, W. Y. DeepPath: A reinforcement learning method for knowledge graph reasoning. In Proceedings of the 2017 Conference on Empirical Methods in Natural Language Processing (pp. 564-573) (2017)

\bibitem{Ref20184}
Adam Santoro, Ryan Faulkner, David Raposo, Jack Rae, Mike Chrzanowski, Theophane Weber, Daan Wierstra, Oriol Vinyals, Razvan Pascanu, \& Timothy Lillicrap, Relational recurrent neural networks, Advances in Neural Information Processing Systems, (pp. 7299-7310) (2018)

\bibitem{Ref20185}
Fabio Pardo, Vitaly Levdik, \& Petar Kormushev, Q-map: A convolutional approach for goal-oriented reinforcement learning, arXiv preprint arXiv:1810.02927 (2018)


\bibitem{Ref20177}
Achiam, J., \& Sastry, S., Surprise-based intrinsic motivation for deep reinforcement learning, arXiv preprint arXiv:1703.01732 (2017)

\bibitem{Ref20172}
Azar, M. G., Osband, I., \& Munos, R., Minimax regret bounds for reinforcement learning, In Proceedings of the 34th International Conference on Machine Learning, (pp. 263-272) (2017)

\bibitem{Ref20176}
Bellemare, M. G., Dabney, W., \& Munos, R., A distributional perspective on reinforcement learning, In Proceedings of the 34th International Conference on Machine Learning, (pp. 449-458) (2017)

\bibitem{Ref20162}
Bellemare, M., Srinivasan, S., Ostrovski, G., Schaul, T., Saxton, D., \& Munos, R., Unifying count-based exploration and intrinsic motivation, In Proceedings of the 29th Annual Conference on Neural Information Processing Systems, (pp. 1471-1479) (2016)


\bibitem{Ref20164}
Brockman, G., Cheung, V., Pettersson, L., Schneider, J., Schulman, J., Tang, J., \& Zaremba, W., OpenAI gym, arXiv preprint arXiv:1606.01540 (2016)

\bibitem{Ref20173}
Fortunato, M., Azar, M. G., Piot, B., Menick, J., Osband, I., Graves, A., Mnih, V., Munos, R., Hassabis, D., Pietquin, O., Legg, S., \& Blundell, C., Noisy networks for exploration, In Proceedings of the 6th International Conference on Learning Representations (Poster) (2018)

\bibitem{Ref20181}
Hessel, M., Modayil, J., van Hasselt, H., Schaul, T., Ostrovski, G., Dabney, W., Horgan, D., Piot, B., Azar, M., \& Silver, D., Rainbow: Combining improvements in deep reinforcement learning, In Proceedings of the 32nd AAAI Conference on Artificial Intelligence, (pp. 3215-3222) (2018)

\bibitem{Ref20161}
Houthooft, R., Chen, X., Duan, Y., Schulman, J., De Turck, F., \& Abbeel, P., Vime: Variational information maximizing exploration, In Proceedings of the 29th Annual Conference on Neural Information Processing Systems, (pp. 1109-1117) (2016)

\bibitem{Ref20141}
Kingma, D. P., \& Ba, J., Adam: A method for stochastic optimization, In Proceedings of the 3rd International Conference on Learning Representations, (2015)

\bibitem{Ref20154}
Lillicrap, T. P., Hunt, J. J., Pritzel, A., Heess, N., Erez, T., Tassa, Y., Silver, D., \& Wierstra, D., Continuous control with deep reinforcement learning, In Proceedings of the 4th International Conference on Learning Representations (2016)

\bibitem{Ref20163}
Mnih, V., Badia, A. P., Mirza, M., Graves, A., Lillicrap, T., Harley, T., Lillicrap, T. P., Silver, D., \& Kavukcuoglu, K., Asynchronous methods for deep reinforcement learning, In Proceedings of the 33rd International Conference on Machine Learning,  (pp. 1928-1937) (2016)

\bibitem{Ref20151}
Mnih, V., Kavukcuoglu, K., Silver, D., Rusu, A. A., Veness, J., Bellemare, M. G., Graves, A., Riedmiller, M., Fidjeland, A. K., Ostrovski, G., Petersen, S., Beattie, C., Sadik, A., Antonoglou, I., King, H., Kumaran, D., Wierstra, D., Legg, S., \& Hassabis, D., Human-level control through deep reinforcement learning, Nature, 518(7540), 529 (2015)

\bibitem{Ref20165}
Osband, I., Blundell, C., Pritzel, A., \& van Roy, B., Deep exploration via bootstrapped DQN, In Proceedings of the 29th Annual Conference on Neural Information Processing Systems, (pp. 4026-4034) (2016)

\bibitem{Ref20182}
Pardo, F., Levdik, V., \& Kormushev, P., Goal-oriented trajectories for efficient exploration, arXiv preprint arXiv:1807.02078 (2018)

\bibitem{Ref20171}
Plappert, M., Houthooft, R., Dhariwal, P., Sidor, S., Chen, R. Y., Chen, X., Asfour, T., Abbeel, P., \& Andrychowicz, M., Parameter space noise for exploration, In Proceedings of the 6th International Conference on Learning Representations (2018)

\bibitem{Ref20175}
Salimans, T., Ho, J., Chen, X., Sidor, S., \& Sutskever, I., Evolution strategies as a scalable alternative to reinforcement learning, arXiv preprint arXiv:1703.03864 (2017)

\bibitem{Ref20155}
Schulman, J., Levine, S., Abbeel, P., Jordan, M. I., \& Moritz, P., Trust region policy optimization, In Proceedings of the 32nd International Conference on Machine Learning, (pp. 1889-1897) (2015)

\bibitem{Ref20153}
Schaul, T., Quan, J., Antonoglou, I., \& Silver, D. , Prioritized experience replay, In Proceedings of the 4th International Conference on Learning Representations (2016)

\bibitem{Ref1998}
Sutton, R. S., \& Barto, A. G., Introduction to reinforcement learning, Cambridge: MIT Press (1998)

\bibitem{Ref20174}
Tang, H., Houthooft, R., Foote, D., Stooke, A., Chen, O. X., Duan, Y., Schulman, J.,  Turck, D. F., \& Abbeel, P. \# Exploration: A study of count-based exploration for deep reinforcement learning. In Proceedings of the 30th Annual Conference on Neural Information Processing Systems, (pp. 2753-2762) (2017)

\bibitem{Ref20152}
Wang, Z., Schaul, T., Hessel, M., van Hasselt, H., Lanctot, M., \& De Freitas, N., Dueling network architectures for deep reinforcement learning, In Proceedings of the 33nd International Conference on Machine Learning (2016)

\end{thebibliography}
\end{document}